%% file: main.tex
\documentclass[conference,a4paper]{IEEEtran}
\usepackage[utf8]{inputenc} 
\usepackage[T1]{fontenc}
\usepackage{url}              
\usepackage{algorithm}
\usepackage{hyperref}
\usepackage{multirow}
\usepackage{bbding}
\usepackage{color}
\usepackage{amsmath,amssymb,amsfonts}
\usepackage{amsthm}
\usepackage{graphicx}
\usepackage{setspace}
\usepackage{multicol}
\usepackage{multirow}
\usepackage{cite}
\usepackage{float}
\usepackage{color}
\usepackage{alltt, listings}
\usepackage{subfigure}
\usepackage{algpseudocode}
\usepackage{wrapfig}
\usepackage{url}
\usepackage{bm}
\usepackage{authblk}

\newtheorem{assumption}{Assumption}
\newtheorem{theorem}{Theorem}
\newtheorem{lemma}{Lemma}

\theoremstyle{definition}
\newtheorem{definition}{Definition}

\begin{document}

\title{Improved Quantization Strategies for Managing Heavy-tailed Gradients in Distributed Learning} 

\author[a]{Guangfeng Yan}
\author[b]{Tan Li}
\author[c]{Yuanzhang Xiao}
\author[a]{Hanxu Hou}
\author[a]{Linqi Song}
\affil[a]{Department of Computer Science, City University of Hong Kong, Hong Kong SAR}
\affil[b]{Department of Computer Science,
The Hang Seng University of Hong Kong, Hong Kong SAR}
\affil[c]{Department of Electrical and Computer Engineering, University of Hawaii at Manoa, United States}
\renewcommand\Authands{ and }
\maketitle

\begin{abstract}
Gradient compression has surfaced as a key technique to address the challenge of communication efficiency in distributed learning. In distributed deep learning, however, it is observed that gradient distributions are heavy-tailed, with outliers significantly influencing the design of compression strategies. Existing parameter quantization methods experience performance degradation when this heavy-tailed feature is ignored. In this paper, we introduce a novel compression scheme specifically engineered for heavy-tailed gradients, which effectively combines gradient truncation with quantization. This scheme is adeptly implemented within a communication-limited distributed Stochastic Gradient Descent (SGD) framework. We consider a general family of heavy-tail gradients that follow a power-law distribution, we aim to minimize the error resulting from quantization, thereby determining optimal values for two critical parameters: the truncation threshold and the quantization density. We provide a theoretical analysis on the convergence error bound under both uniform and non-uniform quantization scenarios. Comparative experiments with other benchmarks demonstrate the effectiveness of our proposed method in managing the heavy-tailed gradients in a distributed learning environment.
\end{abstract}


\input{1.Introduction}

\input{2.Problem}

\input{3.Compressor}

\input{4.Optimal_Parameters}
\input{5.Experiments}
\input{6.Conclusion}
\clearpage
\bibliographystyle{IEEEtran}
\bibliography{mybib}
\clearpage
\input{Appendix}
\end{document}

%% file: 1.Introduction.tex
\section{Introduction}
\label{introduction}
Distributed learning systems, which enable collaborative model training across multiple nodes or devices, have revolutionized the machine learning landscape. However, one of the most significant hurdles these systems face is the communication overhead. In distributed Stochastic Gradient Descent (DSGD)~\cite{{dean2012large},{bekkerman2011scaling}}, a widely adopted algorithm for distributed training, each communication round requires local clients to upload their model parameters to a central server for integration. With the increasing complexity of models and the growth in parameter size, this communication process has become a substantial bottleneck, challenging the practical limits of network bandwidth and efficiency.

To combat this issue, a variety of compression schemes have been introduced. Techniques such as sparsification~\cite{shi2019distributed}, sketching~\cite{rothchild2020fetchsgd}, and quantization~\cite{alistarh2017qsgd} have been explored to reduce the size of the transmitted data. Among these, quantization has gained widespread popularity due to its direct impact on reducing the number of bits required per parameter, making it a strategic fit for environments with limited communication resources. Employing low-bit representations~\cite{banner2019post}, such as 2, 3, or 4 bits for the quantization of model gradients, presents a promising avenue for mitigating the communication load.

\begin{figure}[ht]
\centerline{\includegraphics[width=0.7\linewidth]{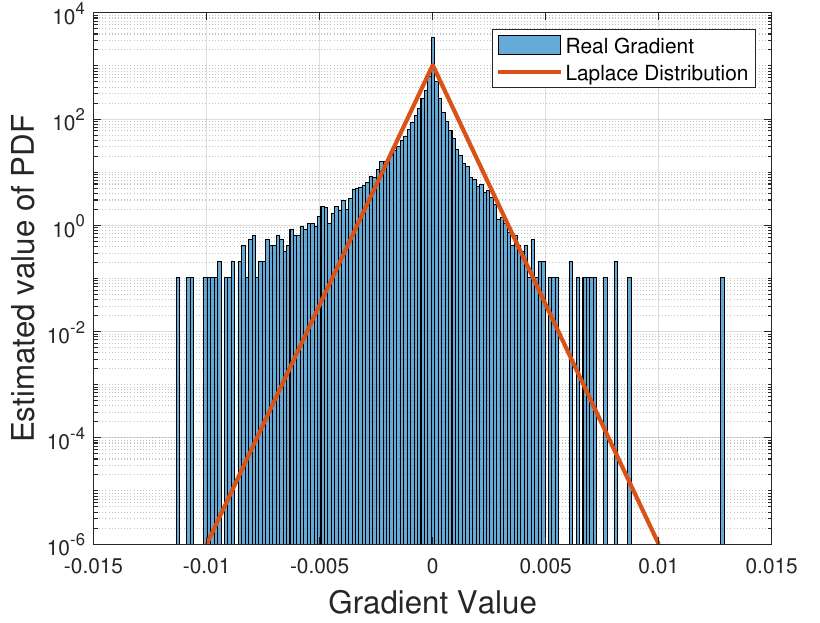}}
	\caption{The probability density of gradient computed with LeNet on MNIST.. (The variance of the Laplace distribution is taken as the same value as the gradient variance.)}
	\label{fig:dist_gradient}
\end{figure}

The effectiveness of gradient quantization techniques often hinges on the assumptions about the statistical distribution of gradients. Previous research efforts~\cite{alistarh2017qsgd,banner2019post} have designed quantization schemes based on assumptions that gradients follow Laplace or Gaussian distributions. However, our empirical analysis of gradients from real-world deep learning models reveals a distinctly heavy-tailed distribution, a crucial detail that has been largely overlooked. Fig.~\ref{fig:dist_gradient} clearly shows that both Laplace and Gaussian distributions exhibit tails that are too `thin' to accurately estimate the true gradient distribution, which is more accurately modeled with `heavier' tails.

Researchers have adopted the Weibull distribution~\cite{liu2023m22} as a more fitting representation of gradient distributions, leading to the design of non-uniform quantization schemes. However, these schemes have not been integrated with truncation techniques, which have been extensively used in communications~\cite{chen2023quantizing}. Gradient truncation is essential for addressing the extreme values within a heavy-tailed distribution, which can significantly skew the quantization process. More importantly, existing quantization works have largely focused on single-client scenarios targeted at improving inference efficiency, rather than on the unique communication requirements of distributed learning environments.

In this paper, we introduce a novel quantization framework specifically designed for distributed learning systems grappling with inherently heavy-tailed gradient distributions. Note that a key difference between quantization in distributed learning and quantization in signal processing for communications is that the goal of quantization in distributed learning is to mitigate the effects on learning convergence, another form of `distortion'. Our main contributions are as follows: 

(1) We propose a two-stage quantizer that initially truncates extreme gradient values before quantization, ensuring that communication cost remains within a pre-determined budget. 

(2) We provide an in-depth analysis of how both truncation and quantization individually and jointly affect the quantization error terms in the convergence error bounds. 

(3) We assume that the tail of the gradient follows a power-law distribution and determine the parameters of the design quantizer, namely, the truncation threshold and quantization density, by minimizing the quantization error.

(4) We validate the effectiveness of our method through theoretical analysis of convergence performance and empirical experiments on actual deep learning models.

%% file: 2.Problem.tex
\section{Problem Formulation}

We consider a distributed learning problem, where $N$ clients collaboratively train a shared model via a central server.
The local dataset located at client $i$ is denoted as $\mathcal{D}^{(i)}$. The objective is to minimize the empirical risk over the data held by all clients, i.e., solve the optimization problem 
\begin{equation}
\begin{array}{llll}	\min_{\bm{\theta} \in \mathbb{R}^d}F(\bm{\theta}) = \sum_{i=1}^N  w_i \mathbb{E}_{\xi^{(i)}\sim \mathcal{D}^{(i)}}[\ell(\bm{\theta};\xi^{(i)})],
	\end{array}\label{optim_problem}
\end{equation}
where $w_i = \frac{|\mathcal{D}^{(i)}|}{\sum_{i=1}^N|\mathcal{D}^{(i)}|}$ is the weight of client $i$, $\xi^{(i)}$ is randomly sampled from $\mathcal{D}^{(i)}$ and $\ell(\bm{\theta};\xi^{(i)})$ is the local loss function of the model $\bm{\theta}$ towards data sample $\xi^{(i)}$. A standard approach to solve this problem is DSGD~\cite{{yan2022ac},{bekkerman2011scaling}}, where each client $i$ first downloads the global model $\bm{\theta}_t$ from server at iteration $t$, then randomly selects a batch of samples $B^{(i)}_t \subseteq D^{(i)}$ with size $B$ to compute local stochastic gradient with $\bm{\theta}_t$: $\bm{g}^{(i)}_t = \frac{1}{B} \sum_{\xi^{(i)} \in B^{(i)}_t} \nabla \ell(\bm{\theta}_t;\xi^{(i)})$. Then the server aggregates these gradients and updates: $\bm{\theta}_{t+1} = \bm{\theta}_t - \eta\sum_{i=1}^N w_i \bm{g}^{(i)}_t$, where $\eta$ is the server learning rate. We make the following two common assumptions on the raw gradient $\nabla \ell(\bm{\theta}_t;\xi^{(i)})$ and the objective function $F(\bm{\theta})$~\cite{{bottou2018optimization},{data2023byzantine}}:
\begin{assumption}[Bounded Variance]
	For parameter $\bm{\theta}_t$, the stochastic gradient $\nabla \ell(\bm{\theta}_t;\xi^{(i)})$ sampled from any local dataset have uniformly bounded variance for all clients:
	\begin{align}
	\mathbb{E}_{\xi^{(i)}\sim \mathcal{D}^{(i)}}\left[\|\nabla \ell(\bm{\theta}_t;\xi^{(i)})-\nabla F(\bm{\theta}_t)\|^2\right] \le \sigma^2.
	\end{align}
	\label{ass:stochastic_gradient} 
\end{assumption}
\vspace{-5mm}
\begin{assumption}[Smoothness]
	The objective function $F(\bm{\theta})$ is $\nu$-smooth: $\forall \bm{\theta},\bm{\theta}' \in \mathbb{R}^d$, $\|\nabla F(\bm{\theta})-\nabla F(\bm{\theta}')\| \leq \nu\|\bm{\theta}-\bm{\theta}'\|$.
	\label{ass:smoothnesee} 
\end{assumption}
To reduce the communication cost, we consider to compress the local stochastic gradients $\bm{g}^{(i)}_t$ before sending them to the server: $\bm{\theta}_{t+1} = \bm{\theta}_t - \eta\sum_{i=1}^N w_i \mathcal{C}_{b}[\bm{g}^{(i)}_t]$,
where $\mathcal{C}_{b}[\cdot]$ is the operator to quantize each element of $\bm{g}^{(i)}_t$ into $b$ bits. Next, we will introduce how to design this quantizer and determine its parameters.

%% file: 3.Compressor.tex
\section{Truncated Quantizer for Heavy-Tail Gradients}
In this section, we introduce a two-stage quantizer designed to address the challenge of heavy-tailed gradients. We begin by detailing the two operative steps, truncation and general stochastic quantization, and analyze their impact on the convergence error. 

\subsection{Two-Stage Quantizer}
In this paper, we form the quantizer using a two-stage operation.


\textbf{Gradient Truncation} The truncation operation cuts off the gradient so that the value is within a range. For an element $g$ of gradient $\bm{g}$, the $\alpha$-truncated operator $\mathcal{T}_{\alpha}[g]$ is defined as 
\begin{equation}
     \mathcal{T}_{\alpha}[g] = \begin{cases} g,& \text{for $|g|\le \alpha$,}\\ 
    \text{sgn}(g)\cdot \alpha,& \text{for $|g|> \alpha$} \end{cases}
 \label{eq:truncated_operation}
\end{equation} 
where $\alpha>0$ is a truncation threshold that determines the range of gradients, and $\text{sgn}(g)\in\{+1,-1\}$ is the sign of $g$. A common intuition is that the thicker the tail of the gradient distribution, the larger the value of $\alpha$ should be set to ensure that the discarded gradient information is upper bounded.

\textbf{Gradient Quantization} For the post-truncation gradient, we propose a general stochastic quantization scheme in an element-wise way. To clarify, consider a truncated gradient element $g$ that falls within the interval $[a_1,a_2]$. To satisfy communication constraints, we aim to encode it using $b$ bits. This encoding process results in $2^b$ discrete quantization points, which effectively divide the interval $[a_1,a_2]$ into $s=2^b-1$ disjoint intervals. The boundaries of these intervals are defined by the points $a_1=l_0 < l_1 \ldots < l_s=a_2$. Each $k$-th interval is denoted by $\Delta_k \triangleq [l_{k-1}, l_k]$, and has a length (or a quantization step size) of $|\Delta_k| = l_k - l_{k-1}$. If $g \in \Delta_k$, we have
\begin{equation}\
	\mathcal{Q}[g] = \begin{cases} l_{k-1},& \text{with probability $1-p_r$,}\\ 
	l_k,& \text{with probability $p_r = \cfrac{g-l_{k-1}}{|\Delta_k|}$.} \end{cases}
\label{eq:quantized_operation}
\end{equation}

It is evident that the specific operation of the quantizer depends on the quantization step size $\Delta_k$, which is essentially the coded book $\mathcal{L}\triangleq\{l_0,l_1,...,l_s\}$. This also determines the statistical characteristics of the quantizer, as demonstrated by the following Lemma.
\begin{lemma}[Unbiasness and Bounded Variance]
    For a truncated gradient element $g\in [a_1,a_2]$ with probability density function $p_g(\cdot)$, given the quantization points $\mathcal{L}=\{l_0, l_1,...,l_s\}$, the nonuniform stochastic quantization satisfies:
    \begin{equation}
		\mathbb{E}[\mathcal{Q}[g]] = g
		\label{eq:unbiassness}
	\end{equation}
	and
	\begin{equation}
	  \mathbb{E} \|\mathcal{Q}[g] - g\|^2 \le \sum_{k=1}^{s} \frac{P_k|\Delta_k|^2}{4}
	\label{eq:qsg}
	\end{equation}
	where $P_k = \int_{l_{k-1}}^{l_k}p_g(x)\mathrm{d}x$ and $|\Delta_k|  = l_k - l_{k-1}$.
	\label{lem:qsg}    
\end{lemma}

The complete proof can be found in Appendix~\ref{proof_lemma1}. We further introduce the concept of the `density' of quantization points, defined as $\lambda_s(g) \triangleq \frac{1}{|\Delta(g)|}$. This definition ensures that $\int_{a_1}^{a_2} \lambda_s(g) dg = s$. In the remainder of the paper, we denote a non-uniform quantizer with quantization destiny function $\lambda_s(\cdot)$ by $Q_{\lambda_s}[\cdot]$. By doing this, Lemma~\ref{lem:qsg} can be rewritten as $\mathbb{E}[\mathcal{Q}_{\lambda_s}[g]] = g$ and $\mathbb{E} \|\mathcal{Q}_{\lambda_s}[g] - g\|^2 \le \int_{a_1}^{a_2}\frac{p(g)}{4\lambda_s(g)^2}\mathrm{d}g$.

\begin{figure}[ht]
\centerline{\includegraphics[width=1\linewidth]{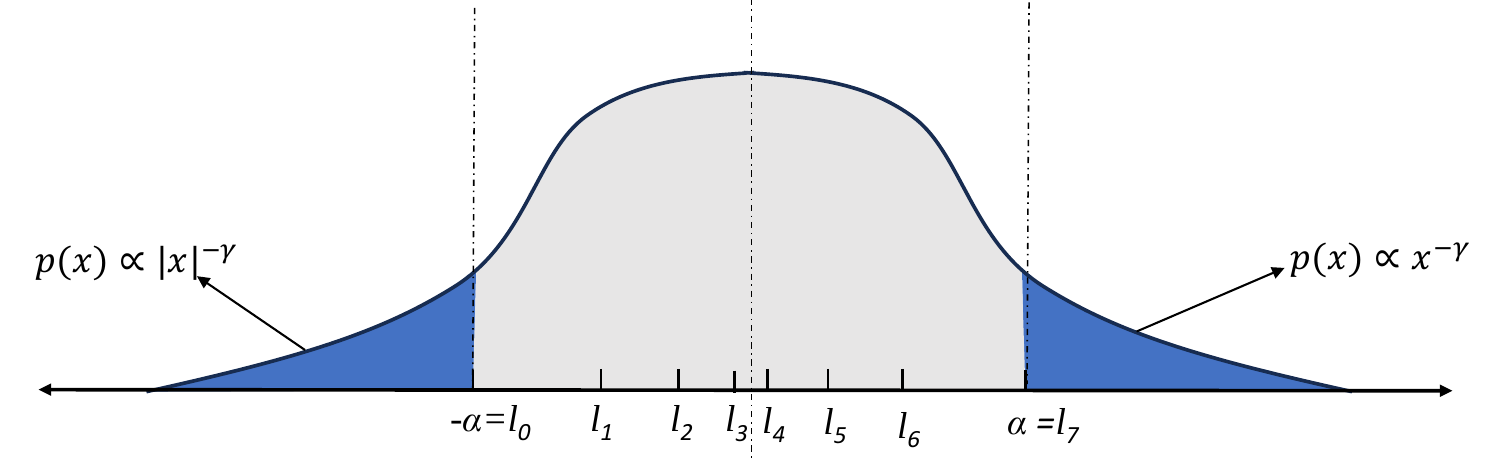}}
	\caption{Two-Stage Quantizer (With truncation threshold $[-\alpha, \alpha]$ and quantization bit $b=3$ and quantization level $s=7$.)}
	\label{fig:Truncated_Quantization}
\end{figure}

To summarize, our proposed two-stage quantizer, denoted as $\mathcal{Q}_{\lambda_s}[\mathcal{T}_{\alpha}(\bm{g})]$, begins with the truncation of gradients $\bm{g}$ using $\mathcal{T}_{\alpha}(\bm{g})$ to curtail values outside the $[-\alpha, \alpha]$ range, thereby reducing the significant gradient noise. These truncated gradients are then quantized through $\mathcal{Q}_{\lambda_s}[\cdot]$ into $b$-bit representations. Please note that, as of now, we have not delineated a specific form for $\lambda_s$. One specific case emerges when $\lambda_s = \frac{s}{2\alpha}$, $\mathcal{Q}_{\lambda_s}[\cdot]$ becomes a uniform quantizer (as seen in the QSGD~\cite{alistarh2017qsgd}), meaning that the truncated range of $2\alpha$, is evenly divided into $s$ intervals. Conversely, when the intervals are not of equal size, $\mathcal{Q}_{\lambda_s}[\cdot]$ becomes a non-uniform quantizer.

For instance, in Fig.~\ref{fig:Truncated_Quantization}, we demonstrate a quantizer that integrates truncation with a non-uniform quantization density. Here, the interval $|l_4-l_3|<|l_1-l_0|$. This is due to a strategy that assigns more quantization points to the peak of the distribution and fewer to the tails. The entire process is encapsulated in the Truncated Quantization for Distributed SGD (TQSGD) algorithm, as shown in Algorithm~\ref{alg:TQSGD}.
\begin{algorithm}[ht] 
	\caption{Truncated Quantizer for Distributed SGD (TQSGD)}
	\begin{algorithmic}[1]
	\State \textbf{Input:} Learning rate $\eta$, initial point $\bm{\theta}_0 \in \mathbb{R}^d$, communication round $T$, parameters of two-stage quantizer $\mathcal{Q}_{\lambda_s}[\mathcal{T}_{\alpha}(\cdot)]$ (truncated threshold $\alpha$, quantization density function $\lambda_s$);
		\For {each communication rounds $t = 0, 1, ..., T-1$:}
		\State \textbf{On each client {$i=1, ..., N$}:}
		\State Download $\bm{\theta}_t $ from server;
		\State Compute the local gradient $\bm{g}^{(i)}_t$ using SGD;
        \State Quantize $\bm{g}^{(i)}_t$ to $\bm{\hat{g}}^{(i)}_t = \mathcal{Q}_{\lambda_s}[\mathcal{T}_{\alpha}(\bm{g}^{(i)}_t)]$ using Eq.~\eqref{eq:truncated_operation} and~\eqref{eq:quantized_operation};
		\State Send $\bm{\hat{g}}^{(i)}_t$ to the server;
 	\State \textbf{On the server:}
        \State Aggregate all quantized gradients  $\bm{\bar g}_t = \sum_{i=1}^N w_i \bm{\hat{g}}^{(i)}_t$;
		\State Update global model parameter: $\bm{\theta}_{t+1} = \bm{\theta}_t - \eta \bm{\bar g}_t$;
	\EndFor
\end{algorithmic} 
	\label{alg:TQSGD}
\end{algorithm}

\subsection{Performance Analysis}
Assuming that each element follows a symmetrical probability density around zero $p(g)$ and is independently and identically distributed, we have the following Lemma to characterize the convergence performance of TQSGD. 

\begin{lemma}
    For a $N$-client distributed learning problem, by applying the two-stage quantizer $\mathcal{Q}_{\lambda_s}[\mathcal{T}_{\alpha}(\cdot)]$ and $w_i = \frac{1}{N}$, the convergence error of Alg.~\ref{alg:TQSGD} for the smooth objective is upper bounded by
	\begin{align} \label{eq:cov_of_TQSGD}
		&\frac{1}{T}\sum_{t=0}^{T-1} \|\nabla F(\bm{\theta}_t)\|^2 \le \underbrace{\frac{2[F(\bm{\theta}_0)-F(\bm{\theta}^*)]}{T\eta}+ \frac{\sigma^2}{NB}}_{\triangleq\mathcal{E}_{DSGD}} \nonumber\\
        &~~+\underbrace{\cfrac{d}{4N}\int_{-\alpha}^{\alpha}\frac{p(g)}{\lambda_s(g)^2}\mathrm{d}g + \cfrac{2d}{N}\int_{\alpha}^{+\infty}(g-\alpha)^2p(g)\mathrm{d}g}_{\triangleq\mathcal{E}_{TQ}} 
	\end{align}
\end{lemma}
This lemma elucidates that the term $\mathcal{E}_{DSGD}$ in Equation \ref{eq:cov_of_TQSGD} delineates the upper limit of the convergence error for the conventional distributed SGD when it is executed with non-compressed model updates. Meanwhile, the second term, $\mathcal{E}_{TQ}$, quantifies the error generated by introducing our two-stage quantization method, reflecting how our algorithm trades off between compression intensity and computational accuracy. The error term $\mathcal{E}_{TQ}$ can be decomposed into two distinct elements: the variance due to quantization (first element) and the bias resulting from truncation (second element). It's crucial to recognize that a minimal truncation threshold $\alpha$ ensures a high density of quantization points as depicted by $\lambda_s(x)$, which effectively reduces the quantization variance towards zero while increasing the truncation bias. In contrast, an elevated threshold $\alpha$ lessens the truncation bias to near zero, yet it escalates the quantization variance. Moreover, the specific distribution of the quantization points $\lambda_s(g)$ plays a pivotal role in dictating the level of quantization variance, thereby affecting the aggregate error term $\mathcal{E}_{TQ}$. For a detailed proof, refer to the Appendix~\ref{proof_lemma2}.

%% file: 4.Optimal_Parameters.tex
\section{Optimal Quantizer Parameter Design}
In this section, we provide theoretical guidance for optimizing the parameters of the proposed quantizer. From the analysis of the process outlined in Alg.~\ref{alg:TQSGD} and its performance, it is evident that determining the parameters of the two-stage quantizer, namely the truncation threshold $\alpha$ and the quantization density function $\lambda_s(g)$, is of critical importance. Formally, we formulate the parameter optimization problem as a \textit{convergence error minimization} problem  under the communication constraints: 
\begin{align}
& \min_{\alpha, \lambda_s} ~ \mathcal{E}_{TQ}(\alpha, \lambda_s) \nonumber\\
& s.t. ~~ \int_{-\alpha}^{\alpha} \lambda_s(x) dx = s
\label{eq:DQP}   
\end{align}

\subsection{Truncated Uniform Quantization}

We first consider a simple case in which $\mathcal{Q}_{\lambda_s}[\cdot]$ operateds as a uniform quantizer~\cite{alistarh2017qsgd}. In this context, the quantization density is constant, $\lambda_s(x) = \frac{s}{2\alpha}$, leading to a coded book composed of evenly spaced points defined as $\mathcal{L}=\{a_1 + k \frac{2\alpha}{s}, k=0,1,...,s\}$. Consequently, our focus shifts to the optimization of $\alpha$. To determine the optimal value of $\alpha$, it is essential to assume the form of $p(g)$, the gradient distribution. In our analysis, we elect to adopt a widely recognized model - the power-law distribution~\cite{clauset2009power}.

\begin{definition}[Power-law distribution~\cite{clauset2009power}]
    A continuous power-law distribution is one described by a probability density $f(x|\gamma, x_{min})$ such that
    \begin{align}\label{eq:power_law}
        f(x|\gamma, x_{min}) = \cfrac{\gamma - 1}{x_{min}^{1-\gamma}} x^{-\gamma}
    \end{align}
    where $x_{min}$ is the lower bound of Power-law distribution, $\gamma$ is the tail index.
\end{definition}

The density function of power-law distribution diverges as $x \to 0$,  so Eq.~\eqref{eq:power_law} cannot hold for all $x \ge 0$; there exists a lower bound $x_{min}$ to the power-law behavior. Hence, in this work, we only use power-law distribution to model the tails of gradients and ignore the intervals near 0, i.e., $\alpha > g_{min}$.
\begin{align}\label{eq:rho_power_law}
    p(g|\gamma, g_{min}, \rho) = \rho\cfrac{\gamma - 1}{g_{min}^{1-\gamma}} |g|^{-\gamma},~~~~\text{for $|g|> g_{min}$}
\end{align}
where $\rho = \int_{g_{min}}^{\infty} p(g)\mathrm{d}g$, and $3<\gamma\le 5$. 
With the power-law distribution, the truncated quantization error in Eq.~\eqref{eq:cov_of_TQSGD} can be rewritten as:
\begin{align}\label{eq:error_TUQ}
   \mathcal{E}_{TQ}(\alpha) &= \cfrac{dQ_U(\alpha)\alpha^2}{Ns^2} + \cfrac{4d\rho g_{min}^{\gamma-1}}{N(\gamma-2)(\gamma-3)} \alpha^{3-\gamma}
\end{align}
where $Q_U(\alpha)\triangleq\int_{-\alpha}^{\alpha}p(g)\mathrm{d}g$. It is difficult to directly solve the above optimization problem. But through alternating iterations, we can obtain approximate numerical results:

\begin{equation}\label{eq:uniform_alpha}
	{\boxed {\alpha = g_{min} \cdot \Big[\cfrac{2\rho s^2}{(\gamma-2)Q_U(\alpha)}\Big]^{\frac{1}{\gamma-1}}}}
\end{equation}
The larger $\gamma$, the thinner the tail of gradients, and the samller the truncation parameter $\alpha$. This aligns with our intuition. We use the following Theorem to characterize the convergence performance of Alg.~\ref{alg:TQSGD} with the Truncated Uniform Quantizer (TQSGD).

\begin{theorem} For an $N$-client distributed learning problem with quantization requirement $s$ , the convergence error of Alg.~\ref{alg:TQSGD} using $\lambda_s(g) = \frac{s}{2\alpha}$ and $\alpha$ in Eq.~\eqref{eq:uniform_alpha} for the smooth objective is upper bounded by
	\begin{align}\label{eq:eq:cov_of_TUQSGD}
		&\frac{1}{T}\sum_{t=0}^{T-1} \|\nabla F(\bm{\theta}_t)\|^2 \le \mathcal{E}_{DSGD}\nonumber\\ 
          &~~~~~~~~+ (\gamma-1)Q_U(\alpha)^{\frac{\gamma-3}{\gamma-1}}\cfrac{d g_{min}^2(2\rho)^{\frac{2}{\gamma-1}}s^{\frac{6-2\gamma}{\gamma-1}}}{N(\gamma-3)(\gamma-2)^{\frac{2}{\gamma-1}}} 
	\end{align}
\end{theorem}

Note that $Q_U(\alpha)\approx 1$. In practice, if we approximate $\alpha$ using $\alpha' \approx g_{min} \cdot \Big[\cfrac{2\rho s^2}{(\gamma-2)}\Big]^{\frac{1}{\gamma-1}}$, the convergence error becomes:
\begin{align}\label{eq:eq:cov_of_TUQSGD2}
    &\frac{1}{T}\sum_{t=0}^{T-1} \|\nabla F(\bm{\theta}_t)\|^2 \le \mathcal{E}_{DSGD}\nonumber\\ 
      &~~~~~~~~+ [(\gamma -3)Q_U(\alpha')+2]\cfrac{d g_{min}^2(2\rho)^{\frac{2}{\gamma-1}}s^{\frac{6-2\gamma}{\gamma-1}}}{N(\gamma-3)(\gamma-2)^{\frac{2}{\gamma-1}}} 
\end{align}
The difference between Eq.\eqref{eq:eq:cov_of_TUQSGD} and Eq.\eqref{eq:eq:cov_of_TUQSGD2} lies in the coefficients of the second term. Specifically, the difference in their coefficients is $\epsilon = (\gamma -3)Q_U(\alpha')+2 - (\gamma-1)Q_U(\alpha)^{\frac{\gamma-3}{\gamma-1}}\le 2[1-Q_U(\alpha')].$ Since $Q_U(\alpha')$ is close to 1, the value of $\epsilon$ is very small. This implies that the difference between the two equations is negligible.

\subsection{Truncated Nonuniform Quantization}
To be more general, we consider that the quantization density can be any positive function. Then the truncated quantization error in Eq.~\eqref{eq:cov_of_TQSGD} can be rewritten as:
\begin{align}\label{eq:error_TNQ}
   \mathcal{E}_{TQ}(\lambda_s(g), \alpha) &= \cfrac{d}{4N}\int_{-\alpha}^{\alpha}\frac{p(g)}{\lambda_s(g)^2}\mathrm{d}g +\cfrac{4d\rho g_{min}^{\gamma-1}}{N(\gamma-3)(\gamma-2)}\alpha^{3-\gamma}
\end{align}

The key question is how to jointly determine $\lambda_s(g)$ and $\alpha$. We use the classic variational principle~\cite{gelfand2000calculus} to construct a Lagrange equation with only $\lambda_s$ as a variable:
\begin{align}
    I(\lambda_s(g), \nu) = \int_{-\alpha}^{\alpha}\Big[\cfrac{p(g)}{\lambda_s(g)^2} - \mu \lambda_s(g)\Big] \mathrm{d}g
\end{align}

According to the Euler-Lagrange equation, we let
\begin{align}
	-\cfrac{2p(g)}{\lambda_s(g)^3}  - \mu = 0
\end{align}

Hence, $\lambda_s(g) = -(\frac{2p(g)}{\mu})^{\frac{1}{3}}$. Using the communication budget constraints Eq.~\eqref{eq:DQP}, we further derive:
\begin{align}\label{eq:nonuiform_lambda}
	{\boxed { \lambda_s(g) = \cfrac{p(g)^{\frac{1}{3}}}{\int_{-\alpha}^{\alpha} p(g)^{\frac{1}{3}} \mathrm{d}g} \cdot s}}
\end{align}
Given fixed values of $s$ and $\alpha$, a larger $p(g)$ necessitates more quantization points for effective compression. For a given gradient distribution $p(g)$ and communication constraint $s$, a larger truncation threshold $\alpha$ means retaining a larger quantization range. But unlike~\cite{panter1951quantization, algazi1966useful}, the above equation is integral to our analysis yet not in $\lambda_s(g)$'s conclusive form until we determine the truncation parameter $\alpha$. We still use the assumption of power-law gradient tail in Eq.~\eqref{eq:rho_power_law}.


Let $Q_N(\alpha) \triangleq \Big[\int_{-\alpha}^{\alpha} p(g)^{\frac{1}{3}} (\frac{1}{2\alpha})^{\frac{2}{3}} \mathrm{d}g \Big]^3$, then we can get the truncation threshold $\alpha$ through alternating iterations:

\begin{equation}\label{eq:nonuiform_alpha}
	{\boxed {\alpha = g_{min} \cdot \Big[\cfrac{2\rho s^2}{(\gamma-2)Q_N(\alpha)}\Big]^{\frac{1}{\gamma-1}}}}
\end{equation}

We use another Theorem to characterize the convergence performance of Alg.~\ref{alg:TQSGD} with Truncated Non-Uniform Quantizer (TNQSGD). 
 
\begin{theorem} For an $N$-client distributed learning problem with quantization requirement $s$, the convergence error of Alg.~\ref{alg:TQSGD} using Eqs.~\eqref{eq:nonuiform_lambda} and~\eqref{eq:nonuiform_alpha} for the smooth objective is upper bounded by
	\begin{align} \label{eq:cov_of_TNQSGD}
		&\frac{1}{T}\sum_{t=0}^{T-1} \|\nabla F(\bm{\theta}_t)\|^2 \le \mathcal{E}_{DSGD}\nonumber\\
  & ~~~~~+ (\gamma-1)Q_N(\alpha)^{\frac{\gamma-3}{\gamma-1}}\cfrac{d g_{min}^2(2\rho)^{\frac{2}{\gamma-1}}s^{\frac{6-2\gamma}{\gamma-1}}}{N(\gamma-3)(\gamma-2)^{\frac{2}{\gamma-1}}}
	\end{align}
\end{theorem}

Using the Holder's inequality, we can get:
\begin{align*}
    Q_N(\alpha)& \triangleq \Big[\int_{-\alpha}^{\alpha} p(g)^{\frac{1}{3}} (\frac{1}{2\alpha})^{\frac{2}{3}} \mathrm{d}g \Big]^3\\
    & \le \Big[(\int_{-\alpha}^{\alpha} p(g)\mathrm{d}g)^{\frac{1}{3}} * (\int_{-\alpha}^{\alpha} \frac{1}{2\alpha}\mathrm{d}g)^{\frac{2}{3}} \Big]^3=Q_U(\alpha)
\end{align*}
This suggests that TNQSGD uses larger truncation threshold $\alpha$, and achieve lower convergence error when compared to TUQSGD.

%% file: 5.Experiments.tex
\section{Experiments}

In this section, we conduct experiments on MNIST to empirically validate our proposed TQSGD and TNQSGD methods. The MNIST consists of 70000 $1 \times 28 \times 28$ grayscale images in 10 classes. We compare our proposed methods with the following baselines: 1) \textbf{QSGD}~\cite{alistarh2017qsgd}, uniform quantization without truncation
; 2) \textbf{NQSGD}, non-uniform quantization without truncation
; 3) oracle \textbf{DSGD}, where clients send non-compressed gradients to the server.

\textbf{Experimental Setting.}  We conduct experiments for $N = 8$ clients and use AlexNet~\cite{krizhevsky2012imagenet} for all clients. We select the momentum SGD as an optimizer, where the learning rate is set to 0.01, the momentum is set to 0.9, and weight decay is set to 0.0005. Note that gradients from convolutional layers and fully-connected layers have different distributions~\cite{wen2017terngrad}.
We thus quantize convolutional layers and fully-connected layers independently. We estimate $\gamma$ based on maximum likelihood estimation: $\gamma = 1 + n \Big[\sum_{j=1}^n \ln{\frac{g_j}{g_{min}}}\Big]^{-1}$, where $g_j, j = 1, . . . , n$, are the gradient values such that, $g_j>g_{min}$.

\begin{figure}
    \centering
    \includegraphics[width=0.75\linewidth]{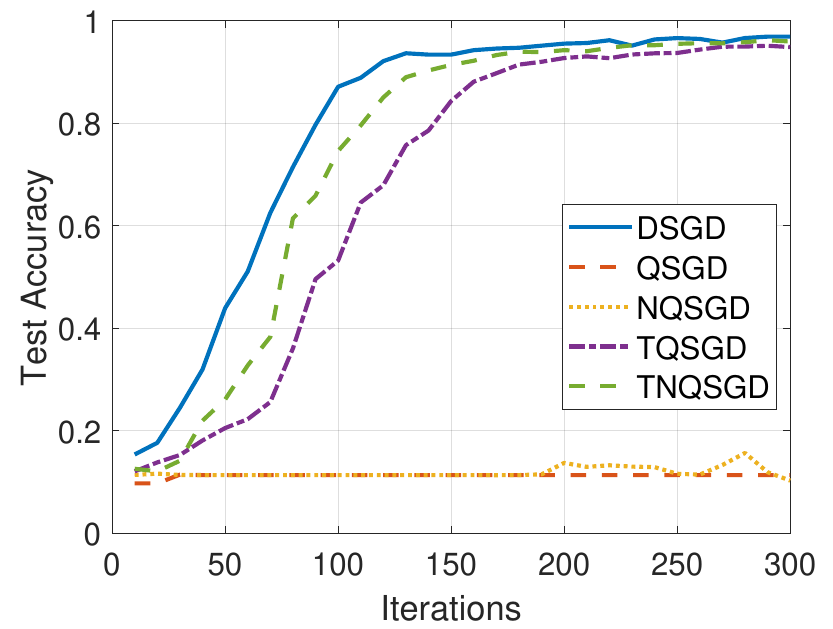}
    \caption{Model performance of different algorithms.}
    \label{fig:testing_performance}
\end{figure}
Fig.~\ref{fig:testing_performance} illustrates the test accuracy of algorithms on MNIST. DSGD achieves a test accuracy of 0.9691 with 32-bit full precision gradients. When $b=3$ bits, TQSGD and TNQSGD achieve test accuracies of 0.9515 and 0.9619, respectively. In contrast, QSGD and NQSGD are almost unable to converge. Results demonstrate that truncation operation can significantly improve the test accuracy of the model under the same communication constraints. Additionally, non-uniform quantization can further enhance the algorithm's performance.

\begin{figure}
    \centering
    \includegraphics[width=0.75\linewidth]{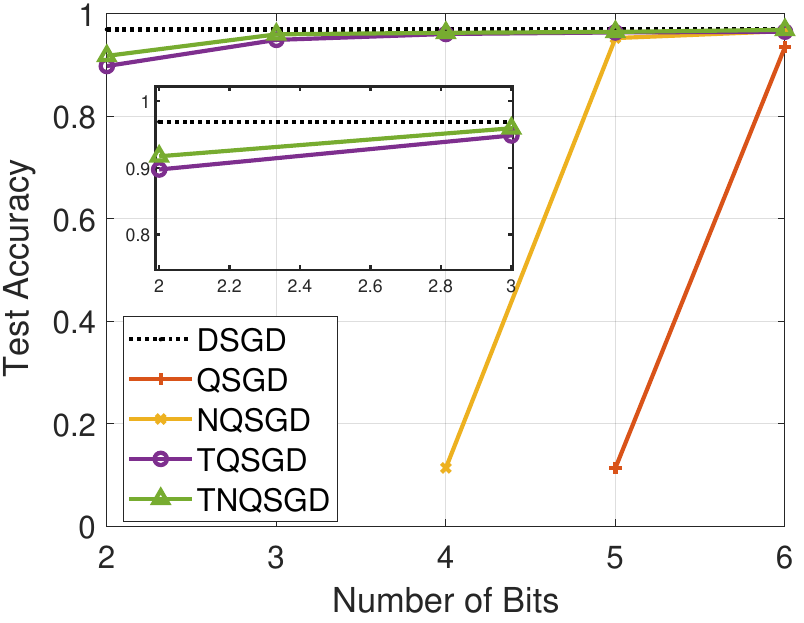}
    \caption{Communication-learning tradeoff of different algorithms.}
    \label{fig:Communication-Learning_tradeoff}
\end{figure}

Fig.~\ref{fig:Communication-Learning_tradeoff} illustrates the tradeoff between communication budget and learning performance in terms of test accuracy of various algorithms. We compare this tradeoff between our proposed algorithms and two other baselines - QSGD and NQSGD. Additionally, we list the accuracy achieved by DSGD without communication budget constraints as a benchmark. All three algorithms exhibit a communication-learning tradeoff; that is, the higher the communication budget, the higher the test accuracy. However, our proposed TQSGD and TNQSGD achieve higher test accuracies than the other two under the same communication cost.

%% file: 6.Conclusion.tex
\section{Conclusion}
We addressed the challenge of communication efficiency in distributed learning through a novel gradient quantization scheme designed for heavy-tailed distributions. Our approach combines gradient truncation with quantization tailored for the communication-constrained distributed SGD. By assuming a power-law distribution for gradient tails, we refined the quantization process to minimize error and optimally set key parameters, truncation threshold, and quantization density. Empirical results demonstrated that our method effectively manages heavy-tailed gradients, outperforming existing benchmarks in distributed settings. 

%% file: Appendix.tex
\section{Appendix}
\subsection{Proof of Lemma 1}
\label{proof_lemma1}
If $x \in \Delta_k$, we have
\begin{align*}\
	\mathcal{Q}[x] = \begin{cases} l_{k-1},& \text{with probability $1-p_r$,}\\ 
	l_k,& \text{with probability $p_r = \cfrac{x-l_{k-1}}{|\Delta_k|}$.} \end{cases}
\end{align*} 

Hence,

\begin{align*}
	 &~~~~~\mathbb{E} \|\mathcal{Q}[x] - x\|^2\\
  &= \sum_{k=1}^{s} \int_{l_{k-1}}^{l_k} \Big[ (l_{k-1}-x)^2\cfrac{l_k-x}{|\Delta_k|} + (l_k-x)^2\cfrac{x-l_{k-1}}{|\Delta_k|}\Big]p(x)\mathrm{d}x\\
  &= \sum_{k=1}^{s} \int_{l_{k-1}}^{l_k} (x-l_{k-1})(l_k-x)p(x)\mathrm{d}x\\
  &\overset{(a)}{\approx}\sum_{k=1}^{s} \int_{l_{k-1}}^{l_k} (x-l_{k-1})(l_k-x) \cfrac{P_k}{|\Delta_k|}\mathrm{d}x\\
  & = \sum_{k=1}^{s} \cfrac{P_k}{|\Delta_k|} \cfrac{(l_k - l_{k-1})^3}{6}\\
  & = \sum_{k=1}^{s} P_k \cfrac{|\Delta_k|^2}{6}
\end{align*}

where $(a)$ uses the high-rate
regime assumption, and $P_k = \int_{l_{k-1}}^{l_k}p(x)\mathrm{d}x$.

\begin{align*}
	 &~~~~~\mathbb{E} \|\mathcal{Q}[x] - x\|^2\\
  &= \sum_{k=1}^{s} \int_{l_{k-1}}^{l_k} \Big[ (l_{k-1}-x)^2\cfrac{l_k-x}{|\Delta_k|} + (l_k-x)^2\cfrac{x-l_{k-1}}{|\Delta_k|}\Big]p(x)\mathrm{d}x\\
  &= \sum_{k=1}^{s}|\Delta_k|^2 \int_{l_{k-1}}^{l_k} \Big[ \cfrac{l_k-x}{|\Delta_k|}\cdot\cfrac{x-l_{k-1}}{|\Delta_k|} \Big]p(x)\mathrm{d}x\\
  &\overset{(b)}{\le} \sum_{k=1}^{s}|\Delta_k|^2 \int_{l_{k-1}}^{l_k} \cfrac{p(x)}{4}\mathrm{d}x\\
  & = \sum_{k=1}^{s} P_k \cfrac{|\Delta_k|^2}{4}
\end{align*}
where $(b)$ uses $y(1-y)\le\cfrac{1}{4}$ for all $y\in[0,1]$.

\subsection{Proof of Lemma 2}
\label{proof_lemma2}
Firstly, we can decompose the mean squared error of the compressed gradient $\mathcal{Q}_{\lambda_s}[ \mathcal{T}_{\alpha}(\bm{g})]$ into a variance term (due to the
nonuniform quantization) and a bias term (due to the truncated operation):

\begin{align}\label{eq:bias_varance}
    \mathbb{E}[\|\mathcal{Q}_{\lambda_s}[ \mathcal{T}_{\alpha}(\bm{g})] - \bm{g}\|^2] &= \underbrace{d\int_{-\alpha}^{\alpha}\frac{p(g)}{4\lambda_s(g)^2}\mathrm{d}g}_{\text{\rm Quantization Variance}}\nonumber\\
    &~~+ \underbrace{2d\int_{\alpha}^{+\infty}(g-\alpha)^2p(g)\mathrm{d}g}_{\text{\rm Truncation Bias}},
\end{align}
 
Using the Assumption~\ref{ass:stochastic_gradient} and Eq.~\eqref{eq:bias_varance}, we have
\begin{align}\label{eq:error_of_aggregated_g}
    &~~~~~\mathbb{E}[\|\bm{\bar g}_t - \nabla F(\bm{\theta}_t)\|^2]\nonumber\\
    &= \cfrac{\sigma^2}{BN} + \cfrac{d}{N}\int_{-\alpha}^{\alpha}\frac{p(g)}{4\lambda_s(g)^2}\mathrm{d}g + \cfrac{2d}{N}\int_{\alpha}^{+\infty}(g-\alpha)^2p(g)\mathrm{d}g
\end{align}

Assumption~\ref{ass:smoothnesee} further implies that $\forall \bm{\theta},\bm{\theta}' \in \mathbb{R}^d$, we have \begin{equation}
F(\bm{\theta}') \leq F(\bm{\theta}) + \nabla F(\bm{\theta})^\mathrm{T} (\bm{\theta}'-\bm{\theta}) + \frac{\nu}{2} \|\bm{\theta}'-\bm{\theta}\|^2.
\label{eq:smooth}
\end{equation}
Hence, we can get
\begin{align*}
F(\bm{\theta}_{t+1}) &\le F(\bm{\theta}_t) + \nabla F(\bm{\theta}_t)^\text{T} (\bm{\theta}_{t+1}-\bm{\theta}_t) + \cfrac{\nu}{2} \|\bm{\theta}_{t+1}-\bm{\theta}_t\|^2\\
&= F(\bm{\theta}_t) - \eta \nabla F(\bm{\theta}_t)^\top \bm{\bar g}_t + \cfrac{\nu\eta^2}{2} \|\bm{\bar g}_t\|^2\\
&\overset{(a)}{\le} F(\bm{\theta}_k) - \eta \nabla F(\bm{\theta}_t)^\top \bm{\bar g}_t + \cfrac{\eta}{2} \|\bm{\bar g}_t\|^2\\
&= F(\bm{\theta}_t) - \cfrac{\eta}{2} \|\nabla F(\bm{\theta}_t)\|^2 + \cfrac{\eta}{2} \|\bm{\bar g}_t-\nabla F(\bm{\theta}_t)\|^2
\end{align*}
where $(a)$ using $\eta \le \cfrac{1}{\nu}$. Then using Eq.~\eqref{eq:error_of_aggregated_g}, we have
\begin{align*}
    \mathbb{E} F(\bm{\theta}_{t+1}) &\le F(\bm{\theta}_t) - \cfrac{\eta}{2} \|\nabla F(\bm{\theta}_t)\|^2 + \cfrac{\eta}{2NB}\sigma^2\\
    & + \cfrac{d\eta}{2N}\int_{-\alpha}^{\alpha}\frac{p(g)}{4\lambda_s(g)^2}\mathrm{d}g + \cfrac{\eta d}{N}\int_{\alpha}^{+\infty}(g-\alpha)^2p(g)\mathrm{d}g
\end{align*}
Applying it recursively, this yields:
\begin{align*}
    &\mathbb{E}[F(\bm{\theta}_T)-F(\bm{\theta}_0)] \le -\cfrac{\eta }{2}\sum_{t=0}^{T-1}\|\nabla F(\bm{\theta}_t)\|^2
    + \cfrac{T\eta}{2NB}\sigma^2\\
    &~~~~~~~ + \cfrac{dT\eta}{2N}\int_{-\alpha}^{\alpha}\frac{p(g)}{4\lambda_s(g)^2}\mathrm{d}g + \cfrac{T\eta d}{N}\int_{\alpha}^{+\infty}(g-\alpha)^2p(g)\mathrm{d}g
\end{align*}

Considering that $F(\bm{\theta}_T) \ge F(\bm{\theta}^*)$, so:
\begin{align}\label{eq:convergence_UQ2}
&\frac{1}{T}\sum_{t=0}^{T-1} \|\nabla F(\bm{\theta}_t)\|^2 \le \cfrac{2[F(\bm{\theta}_0)-F(\bm{\theta}^*)]}{T\eta}+\cfrac{\sigma^2}{NB}\nonumber\\
&~~~~~ + \cfrac{d}{N}\int_{-\alpha}^{\alpha}\frac{p(g)}{4\lambda_s(g)^2}\mathrm{d}g + \cfrac{2d}{N}\int_{\alpha}^{+\infty}(g-\alpha)^2p(g)\mathrm{d}g
\end{align}

\subsection{Proof of Theorem 1}

If we take $\lambda_s(g) = \cfrac{s}{2\alpha}$ and use the assumption of Power-law Gradient Tail, then the truncated quantization error in Eq.~\ref{eq:cov_of_TQSGD} can be rewrite as:
\begin{align*}
   \mathcal{E}_{TQ}(\alpha) &= \cfrac{dQ_U(\alpha)\alpha^2}{Ns^2} + \cfrac{4d\rho g_{min}^{\gamma-1}}{N(\gamma-2)(\gamma-3)} \alpha^{3-\gamma}
\end{align*}

Then we can the truncation threshold $\alpha$ by minimizing above equation:

\begin{equation*}
	\alpha = g_{min} \cdot \Big[\cfrac{2\rho s^2}{(\gamma-2)Q_U(\alpha)}\Big]^{\frac{1}{\gamma-1}}
\end{equation*}
Hence, the truncated quantization error is
\begin{align*}
   \mathcal{E}_{TQ} &= \cfrac{d g_{min}^2(\gamma-1)(2\rho)^{\frac{2}{\gamma-1}}Q_U(\alpha)^{\frac{\gamma-3}{\gamma-1}}}{N(\gamma-3)(\gamma-2)^{\frac{2}{\gamma-1}}} s^{\frac{6-2\gamma}{\gamma-1}}
\end{align*}

Replacing $\cfrac{d}{N}\int_{-\alpha}^{\alpha}\frac{p(g)}{4\lambda_s(g)^2}\mathrm{d}g + \cfrac{2d}{N}\int_{\alpha}^{+\infty}(g-\alpha)^2p(g)\mathrm{d}g$ with $\mathcal{E}_{TQ}$ in Eq.~\eqref{eq:convergence_UQ2}, then we have
\begin{align*}
&\frac{1}{T}\sum_{t=0}^{T-1} \|\nabla F(\bm{\theta}_t)\|^2 \le \cfrac{2[F(\bm{\theta}_0)-F(\bm{\theta}^*)]}{T\eta}+\cfrac{\sigma^2}{N}\\
&~~~~~\cfrac{d g_{min}^2(\gamma-1)(2\rho)^{\frac{2}{\gamma-1}}Q_U(\alpha)^{\frac{\gamma-3}{\gamma-1}}}{N(\gamma-3)(\gamma-2)^{\frac{2}{\gamma-1}}} s^{\frac{6-2\gamma}{\gamma-1}}
\end{align*}


\subsection{Truncated BiScaled Quantization}

\begin{figure}[ht]
\centerline{\includegraphics[width=1\linewidth]{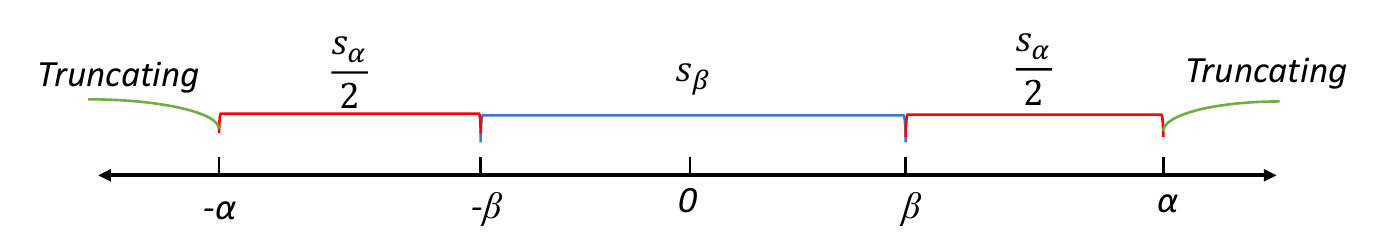}}
	\caption{Truncated BiScaled Quantization.}
	\label{fig:biScaled}
\end{figure}

In this subsection, we consider:
\begin{align}
    \lambda_s(g) = \begin{cases} \cfrac{s_{\beta}}{2\beta},& \text{for $g \in [-\beta, \beta]$,}\\ 
\cfrac{s_{\alpha}}{2(\alpha-\beta)},& \text{for $g \in [-\alpha, -\beta] \cup [\beta,\alpha]$,} \end{cases}
\end{align}
where $s_{\alpha} + s_{\beta} = s = 2^b-1$. And we still use the assumption of power-law gradient tail in Eq.~\eqref{eq:rho_power_law}. Let $p_1 \triangleq \frac{\int_{0}^{\beta} p(g) \mathrm{d}g}{\beta}$ and $p_2 \triangleq \frac{\int_{\beta}^{\alpha} p(g) \mathrm{d}g}{\alpha-\beta}$ denote the average probability density of $g$ in $[0, \beta]$ and $[\beta, \alpha]$, respectively. Then
\begin{align}
    \mathbb{E}[\|\mathcal{Q}_{\lambda_s}[ \mathcal{T}_{\alpha}(\bm{g})] - \bm{g}\|^2] &= \cfrac{2dp_1\beta^3}{s_{\beta}^2} + \cfrac{2dp_2(\alpha-\beta)^3}{s_{\alpha}^2}\nonumber\\
    &~~+ \cfrac{4d\rho g_{min}^{\gamma-1}}{(\gamma-3)(\gamma-2)}\alpha^{3-\gamma},
\end{align}
Then the truncated quantization error in Eq.~\ref{eq:cov_of_TQSGD} can be rewrite as:
\begin{align}
   \mathcal{E}_{TQ}(s_{\alpha}, s_{\beta}, \alpha, \beta) &= \cfrac{2d\bar p_1\beta^3}{Ns_{\beta}^2} + \cfrac{2d \bar p_2(\alpha-\beta)^3}{Ns_{\alpha}^2}\nonumber\\
   &~~~~ +\cfrac{4d\rho g_{min}^{\gamma-1}}{N(\gamma-3)(\gamma-2)}\alpha^{3-\gamma}
\end{align}

The key question is how to determine $s_{\alpha}, s_{\beta}, \alpha$ and $\beta$. To solve that, we formulate it as a convergence error minimization problem under the communication budget constraints:
\begin{align}
    & \min_{s_{\alpha}, s_{\beta}, \alpha, \beta} ~ \mathcal{E}_{TQ} (s_{\alpha}, s_{\beta}, \alpha, \beta) \nonumber\\
    & s.t. ~~~~~~~~~~~~~ s_{\alpha} + s_{\beta}= s,
    \label{eq:optimization-biScaled}
\end{align}
where $s = 2^b - 1$. By solving the above optimization problem, we firstly can determine $s_{\alpha}, s_{\beta}$ as:
\begin{align}\label{eq:s_alpha}
	s_{\alpha}=\cfrac{\bar p_2^{\frac{1}{3}}(1-k)}{\bar p_2^{\frac{1}{3}}(1-k) + \bar p_1^{\frac{1}{3}}k}\cdot s
\end{align}

\begin{align}\label{eq:s_beta}
	s_{\beta}=\cfrac{\bar p_1^{\frac{1}{3}}k}{\bar p_2^{\frac{1}{3}}(1-k) + \bar p_1^{\frac{1}{3}}k}\cdot s
\end{align}
where $k \triangleq \frac{\beta}{\alpha}$. And let $Q_B(\alpha,k) \triangleq \Big[\big(2\int_{k\alpha}^{\alpha}p(g)\mathrm{d}g\big)^{\frac{1}{3}}(1-k)^{\frac{2}{3}} + \big(2\int_0^{k\alpha}p(g)\mathrm{d}g\big)^{\frac{1}{3}}k^{\frac{2}{3}}\Big]^3$, then 
\begin{align}\label{eq:error_TBQ}
    \mathcal{E}_{TQ} (\alpha, k) = \cfrac{dQ_B(\alpha,k)\alpha^2}{Ns^2} +\cfrac{4d\rho g_{min}^{\gamma-1}}{N(\gamma-3)(\gamma-2)}\alpha^{3-\gamma}
\end{align}
Hence, the optimum $k$ and $\alpha$ can be found
for TBQ as follows:
\begin{align}
    (k, \alpha) = \mathop{\arg\min}_k \mathcal{E}_{TQ} (\alpha, k)
\end{align}
Typically, there does not exist a closed solution for above minimization problem. A simple yet effective approximation is to use one step of alternating minimization between $k$ and $\alpha$: $k^* = \mathop{\arg\min}_k Q_B(\alpha,k)$, and
\begin{equation}\label{eq:biscaled_alpha}
	{\boxed {\alpha = g_{min} \cdot \Big[\cfrac{2\rho s^2}{(\gamma-2)Q_B(\alpha,k^*)}\Big]^{\frac{1}{\gamma-1}}}}
\end{equation}

Using Eqs.~\eqref{eq:s_alpha} and~\eqref{eq:s_beta}, we can get the quantization point density function is

\begin{align}\label{eq:biscaled_lamada}
    {\boxed {\lambda_s(g) = \begin{cases} \cfrac{\bar p_1^{\frac{1}{3}}\cdot s}{2\bar p_2^{\frac{1}{3}}(1-k^*)\alpha + 2\bar p_1^{\frac{1}{3}}k^*\alpha},& \text{for $|g|\in [0, k^*\alpha]$,}\\ 
    \cfrac{\bar p_2^{\frac{1}{3}}\cdot s}{2\bar p_2^{\frac{1}{3}}(1-k^*)\alpha + 2\bar p_1^{\frac{1}{3}}k^*\alpha},& \text{for $|g| \in [k^*\alpha,\alpha]$,} \end{cases}}}
\end{align}

If we use Eqs.~\eqref{eq:biscaled_lamada} and~\eqref{eq:biscaled_alpha} to form our truncated quantization in Alg.~\ref{alg:TQSGD}, we can get Truncated BiScaled Quantization for Distributed SGD (TBQSGD). We characterize the convergence performance of TBQSGD in the following Theorem.
\begin{theorem}[Convergence Performance of TBQSGD]
    For an $N$-client distributed learning problem, the quantization bit is $b$, then the convergence error of TBQSGD for the smooth objective is upper bounded by
	\begin{align} \label{eq:cov_of_TBQSGD}
		&\frac{1}{T}\sum_{t=0}^{T-1} \|\nabla F(\bm{\theta}_t)\|^2 \le \mathcal{E}_{DSGD}\nonumber\\
  & ~~~~~+ (\gamma-1)Q_B(\alpha^*,k^*)^{\frac{\gamma-3}{\gamma-1}}\cfrac{d g_{min}^2(2\rho)^{\frac{2}{\gamma-1}}s^{\frac{6-2\gamma}{\gamma-1}}}{N(\gamma-3)(\gamma-2)^{\frac{2}{\gamma-1}}}
	\end{align}
\end{theorem}
Using the Holder's inequality, we can get $Q_B(\alpha^*,k^*) \le 1$.
This suggests that TBQSGD uses larger truncation threshold $\alpha$, and achieve lower convergence error when compared to TUQSGD.


%% file: main.bbl
\begin{thebibliography}{10}
\providecommand{\url}[1]{#1}
\csname url@samestyle\endcsname
\providecommand{\newblock}{\relax}
\providecommand{\bibinfo}[2]{#2}
\providecommand{\BIBentrySTDinterwordspacing}{\spaceskip=0pt\relax}
\providecommand{\BIBentryALTinterwordstretchfactor}{4}
\providecommand{\BIBentryALTinterwordspacing}{\spaceskip=\fontdimen2\font plus
\BIBentryALTinterwordstretchfactor\fontdimen3\font minus \fontdimen4\font\relax}
\providecommand{\BIBforeignlanguage}[2]{{%
\expandafter\ifx\csname l@#1\endcsname\relax
\typeout{** WARNING: IEEEtran.bst: No hyphenation pattern has been}%
\typeout{** loaded for the language `#1'. Using the pattern for}%
\typeout{** the default language instead.}%
\else
\language=\csname l@#1\endcsname
\fi
#2}}
\providecommand{\BIBdecl}{\relax}
\BIBdecl

\bibitem{dean2012large}
J.~Dean, G.~Corrado, R.~Monga, K.~Chen, M.~Devin, M.~Mao, M.~Ranzato, A.~Senior, P.~Tucker, K.~Yang \emph{et~al.}, ``Large scale distributed deep networks,'' in \emph{Advances in Neural Information Processing Systems}, 2012, pp. 1223--1231.

\bibitem{bekkerman2011scaling}
R.~Bekkerman, M.~Bilenko, and J.~Langford, \emph{Scaling up machine learning: Parallel and distributed approaches}.\hskip 1em plus 0.5em minus 0.4em\relax Cambridge University Press, 2011.

\bibitem{shi2019distributed}
S.~Shi, Q.~Wang, K.~Zhao, Z.~Tang, Y.~Wang, X.~Huang, and X.~Chu, ``A distributed synchronous sgd algorithm with global top-k sparsification for low bandwidth networks,'' in \emph{2019 IEEE 39th International Conference on Distributed Computing Systems (ICDCS)}.\hskip 1em plus 0.5em minus 0.4em\relax IEEE, 2019, pp. 2238--2247.

\bibitem{rothchild2020fetchsgd}
D.~Rothchild, A.~Panda, E.~Ullah, N.~Ivkin, I.~Stoica, V.~Braverman, J.~Gonzalez, and R.~Arora, ``Fetchsgd: Communication-efficient federated learning with sketching,'' in \emph{International Conference on Machine Learning}.\hskip 1em plus 0.5em minus 0.4em\relax PMLR, 2020, pp. 8253--8265.

\bibitem{alistarh2017qsgd}
D.~Alistarh, D.~Grubic, J.~Li, R.~Tomioka, and M.~Vojnovic, ``Qsgd: Communication-efficient sgd via gradient quantization and encoding,'' \emph{Advances in Neural Information Processing Systems}, vol.~30, pp. 1709--1720, 2017.

\bibitem{banner2019post}
R.~Banner, Y.~Nahshan, and D.~Soudry, ``Post training 4-bit quantization of convolutional networks for rapid-deployment,'' \emph{Advances in Neural Information Processing Systems}, vol.~32, 2019.

\bibitem{liu2023m22}
Y.~Liu, S.~Rini, S.~Salehkalaibar, and J.~Chen, ``M22: A communication-efficient algorithm for federated learning inspired by rate-distortion,'' \emph{arXiv preprint arXiv:2301.09269}, 2023.

\bibitem{chen2023quantizing}
J.~Chen, M.~K. Ng, and D.~Wang, ``Quantizing heavy-tailed data in statistical estimation:(near) minimax rates, covariate quantization, and uniform recovery,'' \emph{IEEE Transactions on Information Theory}, 2023.

\bibitem{yan2022ac}
G.~Yan, T.~Li, S.-L. Huang, T.~Lan, and L.~Song, ``Ac-sgd: Adaptively compressed sgd for communication-efficient distributed learning,'' \emph{IEEE Journal on Selected Areas in Communications}, vol.~40, no.~9, pp. 2678--2693, 2022.

\bibitem{bottou2018optimization}
L.~Bottou, F.~E. Curtis, and J.~Nocedal, ``Optimization methods for large-scale machine learning,'' \emph{Siam Review}, vol.~60, no.~2, pp. 223--311, 2018.

\bibitem{data2023byzantine}
D.~Data and S.~Diggavi, ``Byzantine-resilient high-dimensional federated learning,'' \emph{IEEE Transactions on Information Theory}, 2023.

\bibitem{clauset2009power}
A.~Clauset, C.~R. Shalizi, and M.~E. Newman, ``Power-law distributions in empirical data,'' \emph{SIAM review}, vol.~51, no.~4, pp. 661--703, 2009.

\bibitem{gelfand2000calculus}
I.~M. Gelfand, R.~A. Silverman \emph{et~al.}, \emph{Calculus of variations}.\hskip 1em plus 0.5em minus 0.4em\relax Courier Corporation, 2000.

\bibitem{panter1951quantization}
P.~Panter and W.~Dite, ``Quantization distortion in pulse-count modulation with nonuniform spacing of levels,'' \emph{Proceedings of the IRE}, vol.~39, no.~1, pp. 44--48, 1951.

\bibitem{algazi1966useful}
V.~Algazi, ``Useful approximations to optimum quantization,'' \emph{IEEE Transactions on Communication Technology}, vol.~14, no.~3, pp. 297--301, 1966.

\bibitem{krizhevsky2012imagenet}
A.~Krizhevsky, I.~Sutskever, and G.~E. Hinton, ``Imagenet classification with deep convolutional neural networks,'' \emph{Advances in neural information processing systems}, vol.~25, 2012.

\bibitem{wen2017terngrad}
W.~Wen, C.~Xu, F.~Yan, C.~Wu, Y.~Wang, Y.~Chen, and H.~Li, ``Terngrad: Ternary gradients to reduce communication in distributed deep learning,'' \emph{Advances in neural information processing systems}, vol.~30, 2017.

\end{thebibliography}
